\PassOptionsToPackage{authoryear,round}{natbib}
\documentclass[preprint,12pt,authoryear]{elsarticle}
\biboptions{authoryear}

\usepackage{amsmath,amssymb}
\usepackage{graphicx}
\usepackage{booktabs}
\usepackage{url}
\usepackage{adjustbox}
\usepackage[linesnumbered,ruled]{algorithm2e}
\usepackage{multirow}
\usepackage{arydshln}
\usepackage{nicematrix}
\usepackage{stfloats}
\usepackage{microtype}
\usepackage{enumitem}
\setlist[itemize]{leftmargin=*, align=parleft, itemsep=0.25ex, topsep=0.6ex, parsep=0pt, partopsep=0pt}
\setlist[enumerate]{leftmargin=*, align=parleft}
\usepackage[hidelinks,breaklinks=true]{hyperref}

\journal{Engineering Applications of Artificial Intelligence}

\newcommand{\R}{\hbox{I \kern -.5em R}}

\begin{document}

\begin{frontmatter}

\title{Discrete Autoregressive Transformer for Generative Mechanism Synthesis}

\author[sbu,cor]{Anar Nurizada\corref{cor1}}
\ead{anar.nurizada@gmail.com}
\cortext[cor1]{Corresponding author}
\author[sbu]{Anurag Purwar}
\address[sbu]{Computer-Aided Design and Innovation Lab, Department of Mechanical Engineering, Stony Brook University, Stony Brook, NY 11794-2300, USA}

\begin{abstract}
\textbf{Engineering application.}
Planar path synthesis requires mechanisms whose coupler curves match a prescribed trajectory; the mapping from curve to linkage is inherently one-to-many across four-, six-, and eight-bar topologies.
We address this design problem with simulation-grounded evaluation on a curated corpus of over one million mechanisms, reporting Chamfer distance and dynamic time warping after forward kinematics and geometric alignment.

\textbf{Artificial intelligence contribution.}
We formulate synthesis as conditional autoregressive sequence modeling: joint coordinates are uniformly quantized to tokens and generated by a decoder-only transformer with a variational-autoencoder (VAE) latent of the target curve and an explicit mechanism-type token.
Training combines token cross-entropy with a Gaussian-smoothed bin auxiliary loss that respects ordinal structure among bins.
At inference, a bounded latent-noise schedule decodes all mechanism types at each noise level; we retain the top five candidates by geometric error, yielding diverse accurate families without dataset lookup.
On held-out tests, aggregate mean Chamfer distance is $0.0132$ and mean dynamic time warping is $0.153$; a latent $k$-nearest-neighbor baseline that conditions on training-set neighbor latents in VAE space achieves matched-topology mean Chamfer distance $0.0071$ and mean dynamic time warping $0.117$ using the same decoder.
\end{abstract}


\end{frontmatter}

\section{Introduction}
\begin{figure*}[htbp]
    \centering
    \includegraphics[width=\linewidth]{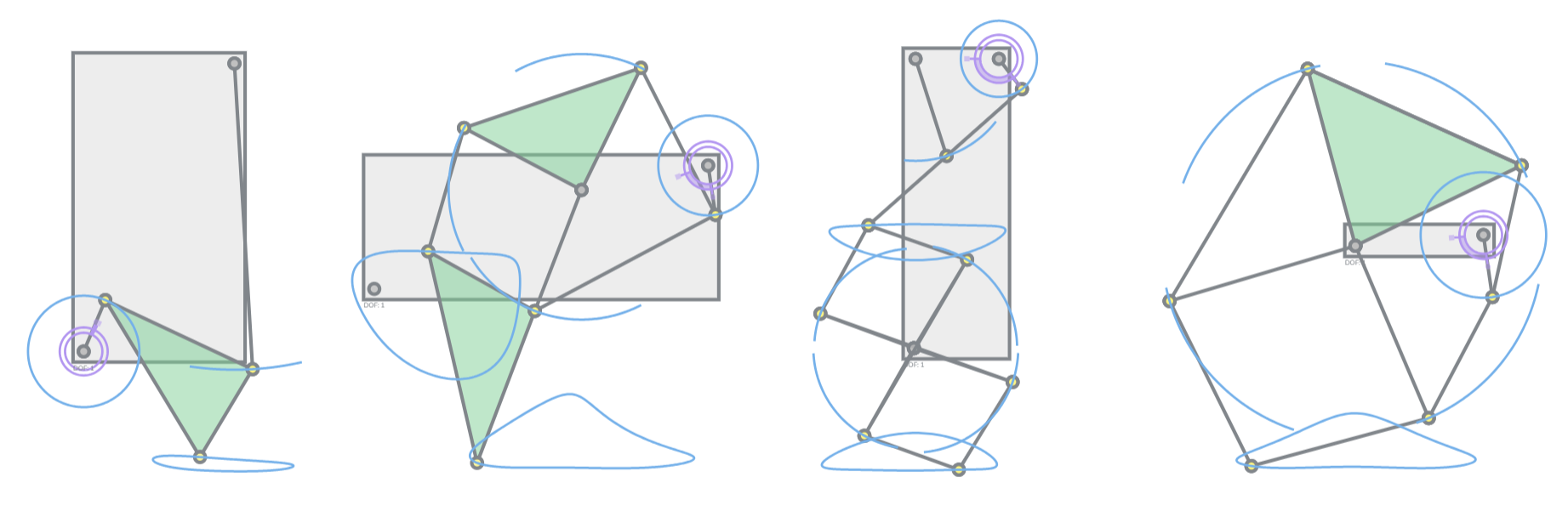}
    \caption{\small Four different types of legged walking mechanisms}
    \label{fig:leg-mechanisms}
\end{figure*}

Mechanism synthesis for path, function, and motion generation has been a central topic in mechanical engineering for decades. In path synthesis, the objective is to design a mechanism whose coupler traces a prescribed trajectory, commonly referred to as a coupler curve. Figure~\ref{fig:leg-mechanisms} shows an example path generation problem where given a coupler curve for walking motion, goal is to synthesize a variety of legged walking mechanisms for integration in robotic systems~\cite{Tang2026}. Classical formulations discretize the desired path into a finite set of precision points, either directly specified or sampled from a continuous path. For example, analytical synthesis methods can yield exact solutions for up to nine precision points for planar four-bar linkage mechanisms~\cite{Wampler1992}. In general, the problem is overdetermined and approximate solutions based on numerical optimization are required~\cite{nocedal2006numerical}.

While optimization-based approaches have been widely adopted, they suffer from several fundamental limitations~\cite{Nurizada2025cBetaVAE}. These methods are often sensitive to initial conditions, computationally expensive, and provide no guarantees of convergence to high-quality solutions. Moreover, they typically ensure interpolation only at discrete precision points, failing to capture the global shape of the desired coupler curve. Objective functions such as the Structural Error, which simultaneously optimize curve shape, scale, orientation, and position~\cite{Erdman2001MechanismDesign}, frequently mischaracterize the synthesis problem and lead to impractical solutions. Perhaps most critically, optimization-based methods generally yield a single mechanism per design task, despite the inherently one-to-many nature of path synthesis. In conceptual mechanism design, however, engineers often seek a diverse set of feasible solutions to accommodate additional constraints such as joint placement, link length ratios, workspace limitations, or manufacturability.

To address some of these challenges, alternative formulations have been proposed that synthesize mechanisms capable of tracing continuous paths rather than interpolating finite point sets. Such approaches explicitly confront the overdetermined nature of coupler-curve synthesis, where the coefficients of the curve equation exceed the available design parameters~\cite{Bai2015}. Although more complex linkages, such as six- and eight-bar mechanisms, offer increased expressive power and can realize intricate motions, their synthesis remains challenging due to the expanded design space and increased computational complexity. Advanced optimization strategies have been applied in this context, yet scalability and robustness remain open problems.

In recent years, approaches grounded in machine learning (ML) have emerged as a promising paradigm for mechanism synthesis. Early work leveraged artificial neural networks (ANNs)~\cite{Hoskins1993, vasiliu2001, Xie2007, galan2009, khan2015, Ahmadi2016, Li2017, Mo2019, Yim2021, Kapsalyamov2023, Yim2023} to learn inverse mappings between coupler curve representations and mechanism parameters, enabling near-instantaneous synthesis once trained. These studies explored a variety of curve representations, including Cartesian coordinates, Fourier descriptors~\cite{khan2015,Li2017}, wavelets~\cite{galan2009}, and image-based encodings~\cite{Nurizada2024InvariantCoupler}. However, most ANN-based methods treat synthesis as a deterministic regression problem and produce only a single mechanism of a fixed type for a given input curve. Nurizada and Purwar~\cite{Nurizada2024InvariantCoupler} systematically demonstrated that different coupler curve representations do not yield meaningful performance differences across learning-based models, highlighting that representation choice alone does not resolve the fundamental limitations of one-to-one prediction.

Subsequent work has increasingly adopted generative formulations to address the one-to-many structure of path synthesis. A comprehensive review by Nurizada and Purwar~\cite{Nurizada2025cBetaVAE} surveyed over three decades of learning-based approaches, including regression models, reinforcement learning~\cite{Vermeer2018,Fogelson2023}, latent-variable generative models~\cite{kingma2014autoencoding}, and hybrid optimization pipelines~\cite{nocedal2006numerical}. Representative recent systems include CLIP-style joint embedding of mechanisms and curves with optimization-based refinement~\cite{nobari2024linklearningjointrepresentations,radford2021learning} and transformer-based sequence modeling over discretized joint coordinates~\cite{Bolanos2025}. Progress in these directions has been enabled by large-scale datasets. The corpus of Nurizada et al.~\cite{Nurizada2025Dataset} contains on the order of three million planar mechanisms across four-, six-, and eight-bar families (including prismatic variants); the LINKS dataset of Nobari et al.~\cite{Nobari2022LINKSDataset} emphasizes richer topological diversity at the cost of many samples being structurally impractical for manufacture. Together, such resources make it realistic to train conditional models that span multiple linkage classes, but they also raise the question of how to evaluate and diversify synthesis under a single, simulation-grounded contract.

Despite this progress, practical gaps remain. Many learning pipelines still emphasize a single predicted mechanism per query, depend on multi-stage architectures, or offer limited control over topology at generation time. Reinforcement-learning approaches~\cite{Vermeer2018,Fogelson2023} can explore large spaces but are often costly per sample, which is awkward for interactive or high-throughput design. There is also a need for inference-time procedures that surface \emph{several} accurate, distinct mechanism families for the same target coupler curve while keeping scores comparable across published methods.

\begin{figure*}[htbp]
    \centering
    \includegraphics[width=\linewidth]{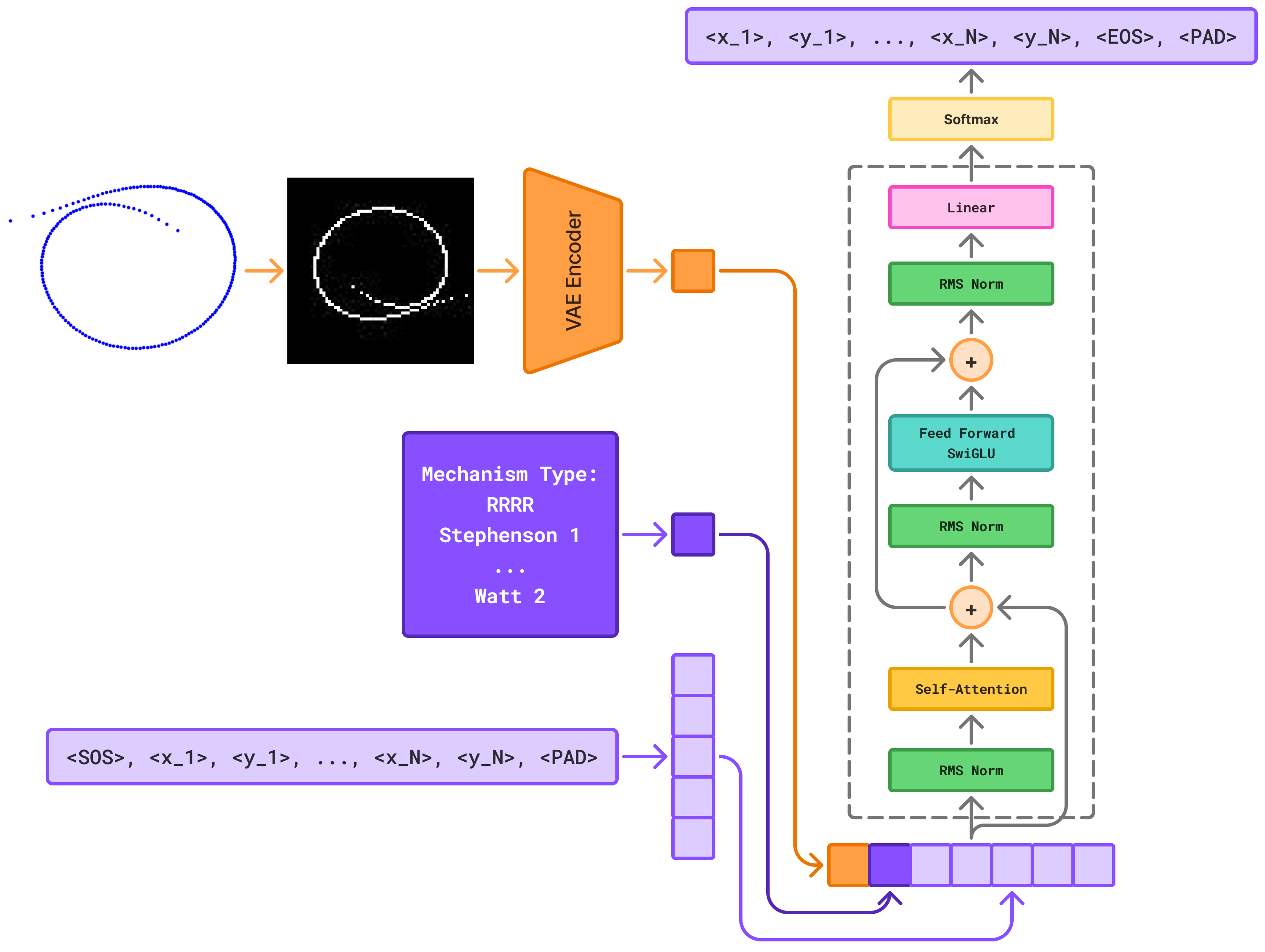}
    \caption{\small Overview of the discrete transformer architecture~\cite{vaswani2017attention,touvron2023llama}. Continuous joint coordinates are discretized into bins and represented as integer tokens~\cite{Bolanos2025,oord2017neural}. A latent geometric token (from the VAE~\cite{kingma2014autoencoding} embedding of the input coupler curve) and a mechanism-type token are prepended to the sequence. A LLaMA-style decoder processes these tokens with causal masking~\cite{vaswani2017attention} and predicts discretized coordinate tokens. Together with the selected mechanism type, the decoded sequence defines a linkage configuration.}
    \label{fig:discrete_model}
\end{figure*}

Motivated by these gaps, we present a single discrete autoregressive transformer on a curated multi-topology subset of~\cite{Nurizada2025Dataset}. The model conditions on a VAE latent of the target coupler curve and on an explicit mechanism-type token, generates quantized joint coordinates with a LLaMA-style causal decoder~\cite{touvron2023llama}, and is trained with token cross-entropy plus a geometry-aware smoothed-bin auxiliary loss~\cite{szegedy2016rethinking}. Figure~\ref{fig:discrete_model} sketches the pipeline; Section~\ref{sec:formulation} fixes notation and data representation, Section~\ref{sec:method} the architecture and training objective, and Section~\ref{sec:results} the simulation-grounded evaluation and inference-time procedures (greedy decoding~\cite{sutskever2014sequence}, latent-noise sweeps with cross-topology top-$K$ selection, and a latent KNN baseline that reuses the same decoder).

Prior learning pipelines differ in how they score coupler agreement: the dataset paper~\cite{Nurizada2025Dataset} illustrates accuracy with bidirectional Chamfer distance from \emph{unsquared} Euclidean nearest-neighbor terms after independent per-curve normalization and a VAE--$k$-NN \emph{retrieval} stage that returns stored candidates, whereas conditional $\beta$-VAE synthesis~\cite{Nurizada2025cBetaVAE} and transformer-based mechanism generation~\cite{Bolanos2025} report dynamic time warping under other normalizations and software conventions. Published scalars are therefore not literal drop-in comparisons to ours. In this work, all reported mechanisms are produced by autoregressive decoding; we do not output dataset mechanism parameters by lookup. The primary protocol conditions on the target curve's VAE latent and on noised variants thereof - points in latent space that need not coincide with any training codevector. A complementary KNN protocol uses the same decoder but conditions on latents taken from training examples near the query in VAE space, so those conditioning vectors were heavily exposed during optimization. We adopt a stringent simulation-grounded protocol - arc-length resampling, centering, a discrete rotation--reflection search that minimizes squared symmetric Chamfer, and banded DTW~\cite{SakoeChiba1978DTW,BerndtClifford1994DTW} - and report state-of-the-art CD/DTW on this multi-topology benchmark relative to these in-protocol evaluations.

The main contributions of this work are:
\begin{itemize}
  \item \textbf{Multi-topology conditional synthesis:} A single autoregressive decoder~\cite{vaswani2017attention,radford2019language} trained on a curated 33-type subset of~\cite{Nurizada2025Dataset} (over $1.2M$ mechanisms), with explicit mechanism-type conditioning for four-, six-, and eight-bar families.
  \item \textbf{Dual conditioning and architecture:} Coupler-curve geometry is injected as a VAE~\cite{kingma2014autoencoding} latent projected to one prefix token~\cite{Nurizada2025Dataset}, together with a learned mechanism-type embedding, feeding a LLaMA-style causal transformer~\cite{touvron2023llama}. Joint coordinates are discretized and generated following~\cite{Bolanos2025,oord2017neural}, as sketched in Fig.~\ref{fig:discrete_model}.
  \item \textbf{Hybrid discrete training objective:} Cross-entropy plus a Gaussian-smoothed bin loss~\cite{szegedy2016rethinking} to respect ordinal structure among quantized coordinates.
  \item \textbf{Simulation-grounded benchmark:} Greedy-decoding evaluation~\cite{sutskever2014sequence} with alignment search prior to CD~\cite{Barrow1977Chamfer} and DTW~\cite{SakoeChiba1978DTW}, together with a latent-space noise sweep~\cite{higgins2017beta} that decodes every mechanism type at multiple perturbed latents and keeps the top five candidates per noise level by (CD, DTW). This procedure delivers the best CD/DTW we report and yields a diverse set of mechanism types among high-scoring solutions for the same target trajectory.
  \item \textbf{Complementary KNN baseline:} The same decoder is evaluated when conditioning on VAE latents of training nearest neighbors~\cite{cover1967nearest} versus the target or noised query latent, highlighting how much error drops when the conditioning point coincides with codes seen repeatedly during training, without any dataset mechanism lookup.
\end{itemize}

The remainder of this paper is organized as follows. Section~\ref{sec:formulation} formulates the synthesis problem and describes the dataset and representations used. Section~\ref{sec:method} details the transformer architecture and training methodology. Section~\ref{sec:results} presents experimental results and discussion. Section~\ref{sec:conclusion} concludes.

\section{Discrete Dataset Representation}\label{sec:formulation}

The mechanisms used in this work are drawn from a large-scale dataset of planar single-degree-of-freedom linkage mechanisms introduced in prior work~\cite{Nurizada2025Dataset}. The full dataset contains approximately three million four-, six-, and eight-bar mechanisms with both open and closed coupler curves, represented using continuous joint coordinates and simulated within a unified kinematic framework. A detailed description of the dataset generation process, normalization procedures, topology encoding, and simulation methodology is provided in~\cite{Nurizada2025Dataset}.

In the present work, a curated subset of this dataset is employed. Specifically, mechanism types with fewer than 20{,}000 samples are excluded in order to reduce class imbalance and avoid over-representation of dominant topologies during training. After this filtering, the resulting dataset comprises 33 distinct mechanism types, including four four-bar mechanism families (with both revolute and prismatic joint variants), sixteen six-bar mechanisms, and thirteen eight-bar mechanisms. This selection preserves substantial topological diversity while yielding a more balanced distribution across mechanism families, facilitating effective conditional generative modeling. There is a total of 1{,}260{,}121 mechanisms in the dataset.

In addition to joint coordinates, each mechanism in the dataset is associated with its corresponding coupler curve representation. In this work, coupler curves are represented as images and encoded using the same variational autoencoder (VAE)~\cite{kingma2014autoencoding} framework introduced alongside the dataset in~\cite{Nurizada2025Dataset}. Specifically, each normalized coupler curve image is mapped to a 50-dimensional latent representation, which captures the global geometric structure of the desired trajectory. This latent vector serves as the geometric conditioning signal for the generative model and is provided as input during both training and inference. By decoupling coupler curve encoding from mechanism generation, this representation enables compact, invariant conditioning while preserving compatibility with the original dataset formulation.

Each planar mechanism is represented using a unified, fixed-length vector of joint and tracer-point coordinates,
\begin{equation}
\mathbf{J} = \big[ (x_1, y_1), (x_2, y_2), \dots, (x_{N_{\max}}, y_{N_{\max}}) \big],
\end{equation}
where $N_{\max}$ denotes the maximum number of joints/tracer points among all mechanism families included in the dataset. Mechanisms with fewer points (e.g., four-bar or six-bar linkages) are padded to length $N_{\max}=12$, using a consistent padding scheme, enabling a single model to operate over multiple topologies while preserving a uniform input/output dimensionality. In addition to $\mathbf{J}$, each mechanism is associated with a discrete mechanism-type label specifying its kinematic topology. This label is provided as an explicit conditioning token during training and generation, allowing the model to synthesize mechanisms across different linkage families while maintaining topology control.

To enable autoregressive sequence modeling, continuous joint coordinates are discretized into a finite set of tokens. Each coordinate dimension is quantized independently over the normalized coordinate range using uniform binning (a choice proposed in~\cite{Bolanos2025}). Specifically, given a normalized coordinate value $x \in [x_{\min}, x_{\max}]$, it is mapped to a discrete bin index
\begin{equation}
q(x) = \left\lfloor \frac{x - x_{\min}}{x_{\max} - x_{\min}} \cdot (K-1) \right\rfloor,
\end{equation}
with $K{=}201$ discrete bins per coordinate dimension.

This transformation converts each joint coordinate into an integer-valued token and each mechanism into an ordered token sequence of fixed length. The discretization resolution is selected to balance representational fidelity and vocabulary size, ensuring that quantization error remains small relative to the scale of the normalized mechanisms while maintaining tractable sequence modeling~\cite{oord2017neural}.

Following discretization, each mechanism is represented as a fixed-length token sequence suitable for autoregressive generation~\cite{vaswani2017attention,radford2019language}. Conditioning information is provided separately and is not part of the predicted output sequence. Specifically, the input to the model is defined as
\begin{equation}
\mathcal{S}_{\text{in}} = [\, z_c,\; z_t,\; \texttt{SOS},\; q(x_1), q(y_1), \dots, q(x_N), q(y_N),\; \dots , \texttt{PAD}],
\end{equation}
where $z_c$ is a latent geometric token encoding the desired coupler curve and $z_t$ is a discrete mechanism-type token (see Fig.~\ref{fig:discrete_model}). The \texttt{SOS} token marks the beginning of the autoregressive generation of joint coordinates. The remaining tokens correspond to discretized joint locations, ordered consistently across all samples. Mechanisms with fewer than $N_{\max}$ joints are padded with \texttt{PAD} tokens to ensure a uniform sequence length.

The training target (label) sequence is defined as
\begin{equation}
\mathcal{S}_{\text{out}} = [\, q(x_1), q(y_1), \dots, q(x_N), q(y_N),\; \texttt{EOS},\; \texttt{PAD}, \dots, \texttt{PAD} \,],
\end{equation}
where \texttt{EOS} denotes the end-of-sequence token. The model is trained to autoregressively predict the joint-coordinate tokens and terminate generation upon emitting \texttt{EOS}, without predicting the conditioning tokens $z_c$ and $z_t$.

This formulation cleanly separates conditioning information from the predicted sequence while allowing the transformer~\cite{vaswani2017attention} to model joint-level and global geometric dependencies through autoregressive decoding~\cite{sutskever2014sequence,radford2019language}.

\section{Discrete Autoregressive Transformer}
\label{sec:method}

This section describes the discrete autoregressive transformer used for generative mechanism synthesis. Mechanism generation is formulated as a conditional sequence prediction problem~\cite{sutskever2014sequence} in which discretized joint coordinates are generated sequentially under geometric and topological constraints.

A decoder-only transformer architecture inspired by autoregressive language models~\cite{radford2019language,vaswani2017attention} is employed. The implementation uses a LLaMA-style configuration
\cite{touvron2023llama} - RMSNorm~\cite{zhang2019root}, SiLU-based feed-forward blocks~\cite{ramachandran2017searching}, and multi-head causal self-attention~\cite{vaswani2017attention} - in a decoder-only stack. The network consists of token embeddings for discretized coordinate values, followed by decoder blocks and a linear output projection. Residual connections~\cite{He2016ResNet} and normalization are applied throughout. Positional embeddings preserve token ordering, and causal masking~\cite{vaswani2017attention} is applied during both training and inference so that each prediction depends only on previously generated tokens. The configuration used for the results reported here has a hidden size of 1{,}536, 32 attention heads, and 6 decoder layers, totaling 229{,}673{,}676 trainable parameters. Training uses distributed data parallel optimization~\cite{paszke2019pytorch} on four NVIDIA A100 GPUs with batch size 512, the AdamW optimizer~\cite{loshchilov2017adamw}, and learning rate $10^{-4}$ (as used for the reported checkpoint). Figure~\ref{fig:discrete_model} summarizes the architecture.

Let $\mathbf{x} = [x_1, x_2, \dots, x_L]$ denote a sequence of discretized joint-coordinate tokens. Under autoregressive decoding, the conditional distribution modeled by the network is
\begin{equation}
p(\mathbf{x} \mid z_c, z_t) = \prod_{i=1}^{L} p(x_i \mid z_c, z_t, x_1, \dots, x_{i-1}),
\label{eq:autoreg-factor}
\end{equation}
where $z_c$ is a latent geometric conditioning token encoding the desired coupler curve and $z_t$ is a discrete mechanism-type token specifying the kinematic topology. Although conditioning tokens are provided as part of the decoder input, they are not included in the predicted output sequence and are excluded from the loss computation.

The latent geometric token $z_c$ is obtained by projecting the 50-dimensional VAE latent through a small multilayer perceptron, producing a single vector in the transformer's embedding space that is treated as one prefix token. The mechanism-type token $z_t$ is embedded using a dedicated lookup table. These two tokens are prepended to the embedded coordinate sequence and serve as global context that conditions all subsequent predictions. Joint-coordinate generation begins only after a dedicated start-of-sequence (\texttt{SOS}) token, ensuring a clear separation between conditioning information and predicted tokens.

We apply a geometry-aware auxiliary loss only at coordinate-token positions $t \in \mathcal{T}_{\text{coord}}$. Let $k^\star_t$ be the ground-truth bin index at $t$, and let $q_t$ be the Gaussian-smoothed target over bins $k$,
\begin{equation}
  q_t(k) \;=\; \frac{\exp\!\left(-\dfrac{(k - k^\star_t)^2}{2\sigma^2}\right)}{\displaystyle\sum_{k'} \exp\!\left(-\dfrac{(k' - k^\star_t)^2}{2\sigma^2}\right)},
  \label{eq:smoothed-target}
\end{equation}
where $\sigma$ is a \emph{smoothing width} (kernel scale) on the bin index; we use $\sigma = 2.0$ in our experiments (a soft-target construction related to label smoothing~\cite{szegedy2016rethinking}). To avoid confusion with softmax temperature used elsewhere, we do not refer to $\sigma$ as a ``temperature.'' The auxiliary term is cross-entropy between $q_t$ and the predicted categorical distribution $p_\theta$,
\begin{equation}
  \mathcal{L}_{\text{geom}}
  \;=\;
  -\sum_{t \in \mathcal{T}_{\text{coord}}} \sum_{k} q_t(k)\,\log p_\theta\!\left(x_t = k \mid z_c, z_t, x_1, \dots, x_{t-1}\right).
  \label{eq:Lgeom-ce}
\end{equation}
Here $p_\theta(x_t \mid z_c, z_t, x_1, \dots, x_{t-1})$ is the same conditional as in~\eqref{eq:autoreg-factor} (with \texttt{SOS} placed after $(z_c,z_t)$ and before $x_1$, as in $\mathcal{S}_{\text{in}}$). The full objective is
\begin{equation}
  \mathcal{L} = \mathcal{L}_{\text{CE}} + \mathcal{L}_{\text{geom}},
  \label{eq:total-loss}
\end{equation}
with unit weight on $\mathcal{L}_{\text{geom}}$ relative to $\mathcal{L}_{\text{CE}}$. Tokens such as \texttt{SOS} and \texttt{EOS} contribute only to $\mathcal{L}_{\text{CE}}$; \texttt{PAD} positions are masked and omitted from both sums.


Training is performed using randomly shuffled mini-batches. During training, ground-truth tokens are provided as input to predict subsequent tokens (teacher forcing~\cite{williams1989learning}). Padding tokens are masked out of the loss, allowing mechanisms of varying complexity to be trained within a unified fixed-length representation. Continuous monitoring metrics include token-level accuracy, discrete bin error, and continuous coordinate error obtained by mapping predicted bins back to bin centers.

At inference time, generation proceeds autoregressively by first providing the conditioning tokens followed by the \texttt{SOS} token. Joint-coordinate tokens are then generated sequentially until an end-of-sequence (\texttt{EOS}) token is emitted or a maximum sequence length is reached, after which padding tokens are appended as needed. The quantitative results in Section~\ref{sec:results} use \emph{greedy} decoding~\cite{sutskever2014sequence}: at each autoregressive step the highest-probability coordinate token is chosen, so the trajectory is deterministic and no randomness is injected. Temperature scaling and stochastic sampling~\cite{radford2019language,touvron2023llama} can be used for exploratory design but were not used for the reported CD/DTW tables.

This formulation enables efficient generation of multiple distinct mechanisms for a single target coupler curve while maintaining explicit control over mechanism topology, thereby capturing the inherently one-to-many structure of the mechanism synthesis problem.

\section{Results}
\label{sec:results}

This section evaluates the discrete autoregressive transformer (Fig.~\ref{fig:discrete_model}). We focus on two questions: whether the model can accurately reconstruct coupler curves under forward kinematic simulation, and whether it supports one-to-many generation across mechanism families for a fixed target trajectory. All results are obtained through simulation-based geometric evaluation; no metrics are computed directly in token space.

Each test sample consists of a target coupler curve and its associated mechanism data from a held-out subset of the dataset described in Section~\ref{sec:formulation}. The coupler curve is first encoded using the pretrained variational autoencoder~\cite{kingma2014autoencoding} introduced in~\cite{Nurizada2025Dataset}, yielding a 50-dimensional latent representation. This latent vector is treated as a fixed geometric descriptor of the desired trajectory and is used as conditioning input for all evaluations involving that curve.

Mechanism generation for the main tables uses deterministic greedy autoregressive decoding~\cite{sutskever2014sequence,vaswani2017attention}. Starting from the \texttt{SOS} token, the model predicts one token at a time by selecting the maximum-probability output at each step, terminating when the \texttt{EOS} token is emitted or when a predefined maximum sequence length is reached. No stochastic sampling, temperature scaling, top-$k$ truncation, or noise injection is used during these evaluations. This design choice ensures that reconstruction performance reflects the learned conditional mapping rather than variability introduced by random sampling.

The generated token sequences are decoded into continuous joint coordinates using the bin centers defined during discretization. These coordinates are then passed to a forward kinematic simulator~\cite{craig2005introduction,Erdman2001MechanismDesign}. Only mechanisms that yield valid simulations with sufficiently long trajectories are retained. From each simulation, the coupler curve corresponding to the appropriate tracer point is extracted for geometric evaluation.

All geometric comparisons are performed after identical preprocessing of the ground-truth and predicted coupler curves. Curves are resampled uniformly by arc length~\cite{docarmo1976differential} to a fixed number of points and centered to remove translational effects. While global normalization removes scale differences, angular normalization alone is not always reliable, as it can result in mirrored or rotated curve configurations that are geometrically equivalent but misaligned under direct comparison.

To resolve this ambiguity, we explicitly search for the best geometric alignment between predicted and target curves. For each predicted curve, a discrete set of in-plane rotations and mirror reflections is evaluated, and the configuration that minimizes the Chamfer Distance (CD)~\cite{Barrow1977Chamfer} with respect to the ground-truth curve is selected. This procedure identifies the appropriate orientation of the predicted curve prior to further evaluation, ensuring that geometric similarity is assessed independently of arbitrary angular or mirroring discrepancies. After the optimal alignment is determined using CD, Dynamic Time Warping (DTW)~\cite{SakoeChiba1978DTW,BerndtClifford1994DTW} is computed between the aligned predicted curve and the ground-truth curve, and the minimum DTW value over forward and reversed traversal directions is reported.

Earlier studies report Chamfer and DTW under different conventions - for example, the dataset manuscript~\cite{Nurizada2025Dataset} uses \emph{unsquared} mean nearest-neighbor Chamfer after per-curve normalization in a VAE--$k$-NN pipeline, while conditional $\beta$-VAE~\cite{Nurizada2025cBetaVAE} and transformer DSL synthesis~\cite{Bolanos2025} use other normalizations and DTW implementations - so raw table values are not identical units to ours. Our scores use squared symmetric Chamfer~\cite{Barrow1977Chamfer,Fan2017Point} after joint $\mathrm{O}(2)$ alignment and banded DTW~\cite{SakoeChiba1978DTW} as specified above. Within this strict simulator-evaluation contract, the aggregate mean CD $0.0132$ and mean DTW $0.153$ (Table~\ref{tab:latent_noise_aggregate}) represent state-of-the-art multi-topology coupler fidelity among the protocols we instantiate, achieved by autoregressive decoding from the target latent and from noised latents along the schedule - not by copying mechanism parameters from the dataset.

Reconstruction accuracy is first evaluated under matched topology conditions. For each test sample, the model is conditioned on the latent representation of the target coupler curve and the mechanism-type token corresponding to the ground-truth mechanism. A single mechanism is generated via greedy decoding, simulated, and compared against the target trajectory. Across the evaluated samples, the model consistently produces mechanisms whose simulated coupler curves closely match the desired geometry, yielding low Chamfer Distance and DTW values in the majority of cases. Visual inspection confirms that the generated curves capture global geometric characteristics such as overall shape, symmetry, and curvature extrema, indicating that discretized autoregressive generation does not impede accurate geometric reconstruction.

The \emph{primary} generative study in this section goes beyond a single greedy decode at the nominal VAE latent~\cite{kingma2014autoencoding,higgins2017beta}. We treat the 50-dimensional latent $\boldsymbol{\mu}$ encoding the target coupler curve as a point in a smooth space and construct a one-parameter family of perturbed latents
\begin{equation}
\tilde{\boldsymbol{z}}(\alpha) = \boldsymbol{\mu} + \alpha\,\sigma\,\boldsymbol{\epsilon},
\end{equation}
where $\alpha$ varies from $0$ to $1$ over a fixed number of steps, $\sigma$ sets the noise scale, and $\boldsymbol{\epsilon}$ is a single draw of standard normal noise per target curve (fixed across $\alpha$), so that increasing $\alpha$ moves along one random direction away from $\boldsymbol{\mu}$. In software, this schedule is implemented in \emph{ray} mode: one pseudorandom direction per test curve (deterministically seeded per sample index) is reused for all $\alpha$. An alternative \emph{step} ablation draws a fresh isotropic Gaussian perturbation at each noise level; all reported latent-noise tables use the ray formulation. At \emph{each} $\alpha$, we condition the decoder on $\tilde{\boldsymbol{z}}(\alpha)$ and, in turn, on \emph{every} mechanism-type token in the label set. Each pair $(\tilde{\boldsymbol{z}}(\alpha),\text{type})$ yields a greedy token sequence, a simulated coupler curve, and the usual CD and DTW after the alignment search described at the beginning of this section. For that noise level we retain only the top five type--decode pairs ranked by CD, with DTW as tie-breaker, discarding poorer candidates. Aggregating over targets and noise levels produces a compact set of high-fidelity mechanisms that differ in topology as well as joint geometry.

This protocol directly targets two goals stressed throughout the paper: (i)~minimize geometric error under simulation, and (ii)~expose diversity in mechanism family among solutions that remain accurate for the same trajectory. In practice the top-$K$ sets exhibit broader coverage of mechanism indices than a single matched-topology greedy decode, while the best CD/DTW values in each sweep meet or improve upon the single-latent greedy baseline. We summarize aggregate CD/DTW over retained predictions and report simple diversity diagnostics - for example, the empirical distribution of mechanism types in the saved top-$K$ sets (entropy~\cite{shannon1948mathematical} relative to a uniform label prior highlights under- or over-represented families). Together, these statistics support the claim that latent-space exploration with cross-topology decoding is an effective inference-time strategy for high-quality, multi-family synthesis.

\begin{figure*}[p]
    \centering
    \includegraphics[width=\textwidth,height=\dimexpr\textheight-6\baselineskip\relax,keepaspectratio]{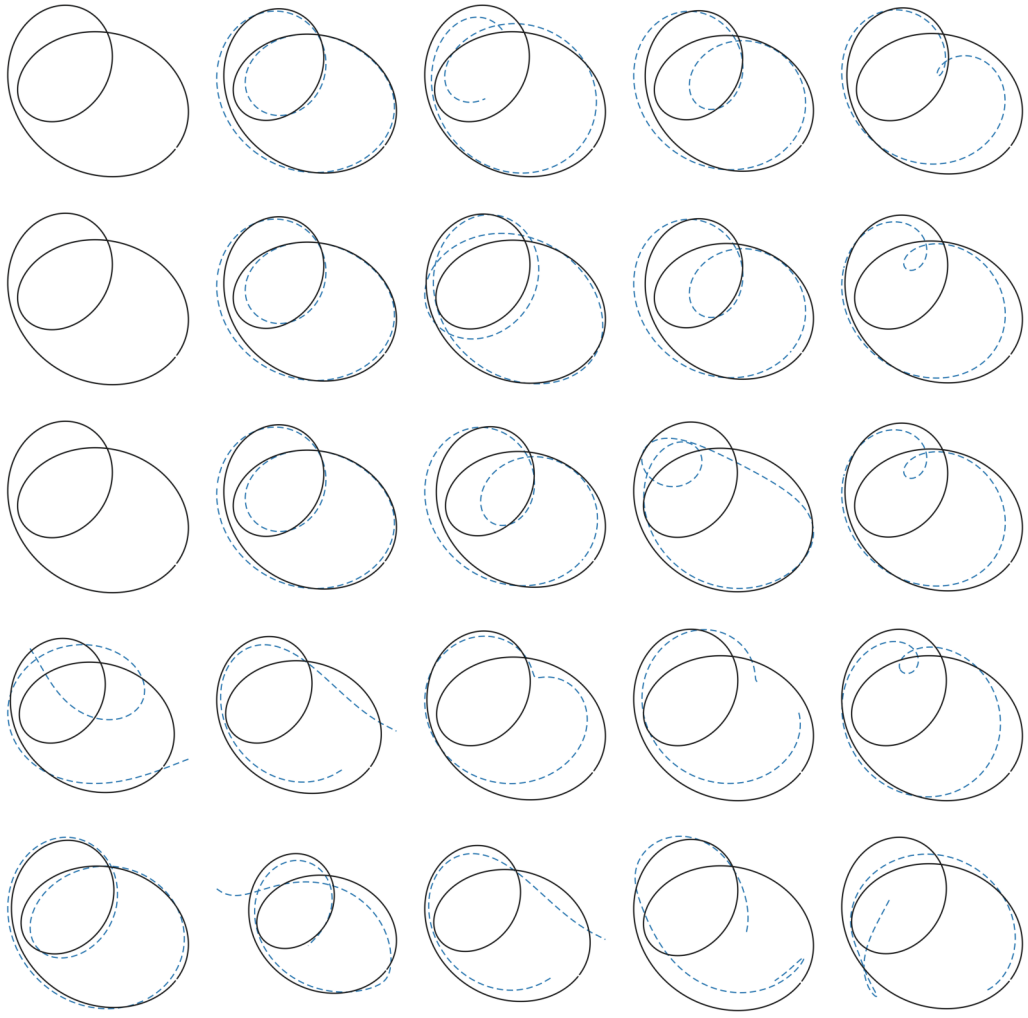}
    \caption{Latent-noise sweep for a target curve. Rows: $\alpha \in \{0, 0.25, 0.5, 0.75, 1\}$ (top to bottom) along a fixed random direction in VAE latent space. Columns: top five mechanisms at that $\alpha$ after cross-topology greedy decoding, simulation, and ranking by Chamfer distance (DTW tie-break). Solid black curve is the desired curve, whereas dotted blue curve is the predicted curve.}
    \label{fig:latent_noise_grid_034}
\end{figure*}

\begin{figure*}[p]
    \centering
    \includegraphics[width=\textwidth,height=\dimexpr\textheight-6\baselineskip\relax,keepaspectratio]{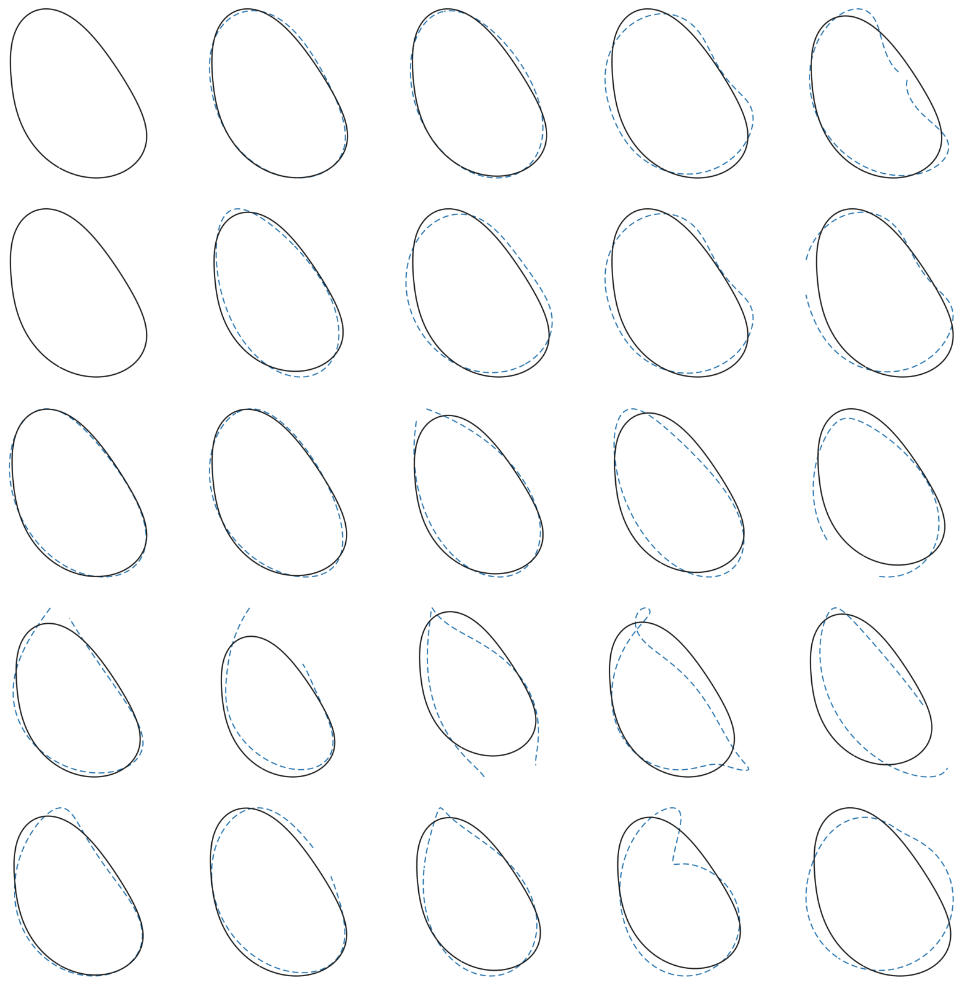}
    \caption{Latent-noise sweep for a target curve; same layout and protocol as Fig.~\ref{fig:latent_noise_grid_034}.}
    \label{fig:latent_noise_grid_079}
\end{figure*}

Table~\ref{tab:latent_noise_gt_mech} reports mean CD and DTW among \emph{matched-topology} rows in the saved top-$K$ lists - that is, evaluations where the mechanism-type token equals the ground-truth type for the target curve - at each noise level $\alpha$ on $100$ held-out samples. The column $n$ counts how often the ground-truth type appears in the top five at that $\alpha$ (it need not appear for every curve once the latent is perturbed). Table~\ref{tab:latent_noise_aggregate} aggregates geometry over all $5{,}000$ saved predictions ($100$ targets $\times$ $10$ noise levels $\times$ $K{=}5$ ranks) and lists diversity of mechanism labels in that pool. Figures~\ref{fig:latent_noise_grid_034} and~\ref{fig:latent_noise_grid_079} visualize the same inference protocol for two representative targets using five evenly spaced noise levels $\alpha \in \{0, 0.25, 0.5, 0.75, 1\}$; each row is a noise level and each column is a rank after sorting candidates by CD (DTW tie-break). Each subplot overlays ground truth (black solid line) and prediction (blue dotted line) after alignment.

\begin{table}[t]
\centering
\caption{Latent-noise sweep, ground-truth mechanism only: mean CD/DTW over saved top-$K$ rows whose evaluated type matches the target, for $100$ test curves per $\alpha$.}
\label{tab:latent_noise_gt_mech}
\begin{tabular}{lccc}
\hline
$\alpha$ & $n$ & Mean CD & Mean DTW \\
\hline
0.0000 & 100 & 0.000113 & 0.000066 \\
0.1111 &  70 & 0.000633 & 0.000445 \\
0.2222 &  50 & 0.002039 & 0.061192 \\
0.3333 &  37 & 0.010728 & 0.068920 \\
0.4444 &  34 & 0.010169 & 0.249816 \\
0.5556 &  27 & 0.012833 & 0.390349 \\
0.6667 &  22 & 0.016132 & 0.319095 \\
0.7778 &  19 & 0.008196 & 0.437816 \\
0.8889 &  21 & 0.018295 & 0.137361 \\
1.0000 &  27 & 0.008788 & 0.131991 \\
\hline
\end{tabular}
\end{table}

\begin{table}[t]
\centering
\caption{Latent-noise sweep with $K{=}5$: aggregate geometry over all saved predictions and label diversity ($33$ mechanism types).}
\label{tab:latent_noise_aggregate}
\begin{tabular}{lc}
\hline
Metric & Value \\
\hline
Mean CD & 0.013218 \\
Median CD & 0.003998 \\
Mean DTW & 0.153457 \\
Median DTW & 0.005949 \\
Distinct types in pool & 33 / 33 \\
\hline
\end{tabular}
\end{table}

For \emph{additional} comparison, we evaluate a latent-space nearest-neighbor protocol~\cite{cover1967nearest} that still uses autoregressive generation~\cite{vaswani2017attention} - not dataset lookup of stored linkages. For each neighbor, the same transformer is conditioned on that neighbor’s VAE latent (a vector that appears in the training set) and on a mechanism-type token, and a mechanism is produced by greedy decoding from scratch. The resulting simulated coupler curve is compared to the query target curve using the same CD/DTW and alignment protocol as elsewhere. Intuitively, neighbor latents lie on the training manifold and were ``seen'' many times during optimization, whereas the primary protocol emphasizes the target latent and directions perturbed away from it (similar to the training distribution but not identical to any stored code). For each test case, the mean and minimum CD and DTW values across the nearest neighbors are recorded.

Using this KNN approach, we report quantitative results under two settings. In both cases, the same set of 100 target coupler curves is used. For each target curve, we take its 30 nearest neighbors in the VAE latent space~\cite{kingma2014autoencoding} (training-set codevectors), yielding 3{,}000 neighbor latents per setting for conditioning the decoder.

In the first setting, \emph{matched-topology KNN evaluation}, each neighbor latent is decoded using only the mechanism type associated with it. This results in $100 \times 30$ total predictions. Table~\ref{tab:knn_matched} summarizes the resulting Chamfer Distance (CD) and Dynamic Time Warping (DTW) statistics. The low mean and median errors indicate that nearby latent codes consistently correspond to mechanisms whose coupler curves closely match the desired geometry.

\begin{figure*}[t]
\centering
\setlength{\tabcolsep}{4pt}%
\begin{tabular}{@{}ccc@{}}
\includegraphics[width=0.31\textwidth]{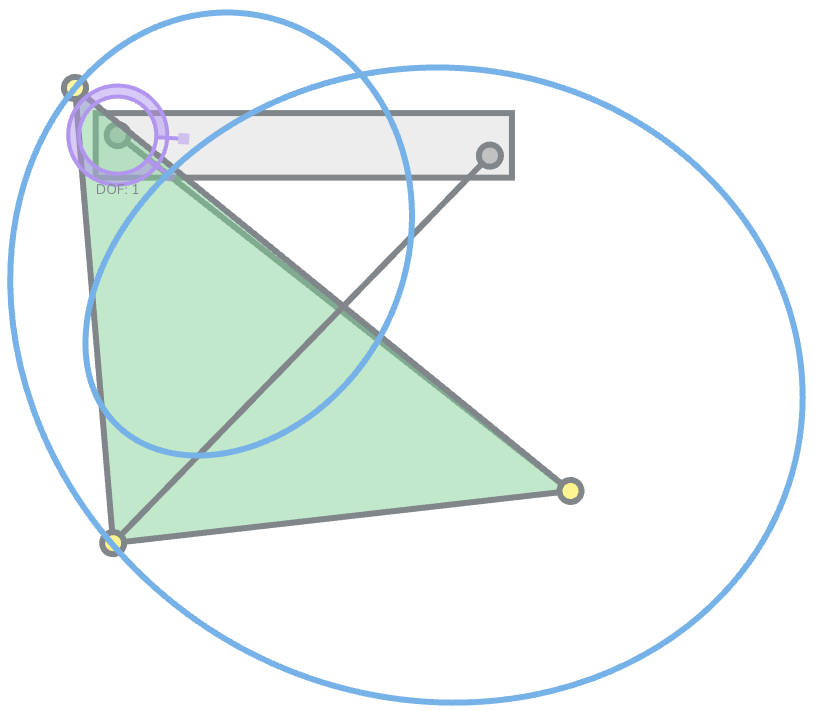} &
\includegraphics[width=0.31\textwidth]{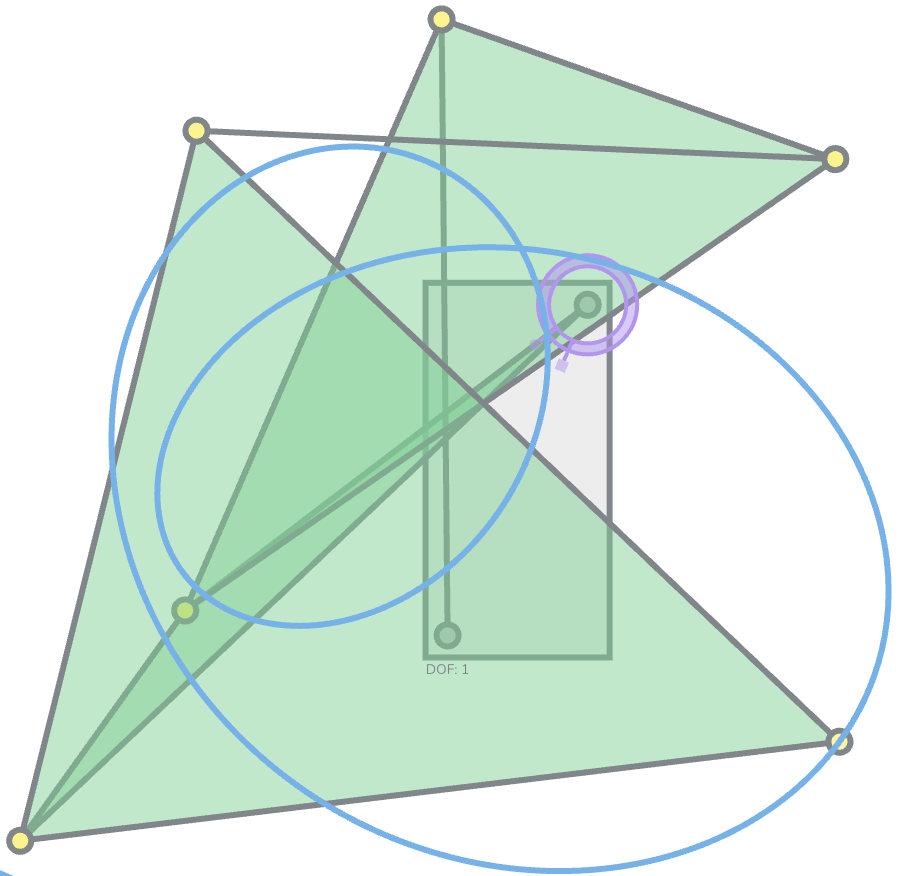} &
\includegraphics[width=0.31\textwidth]{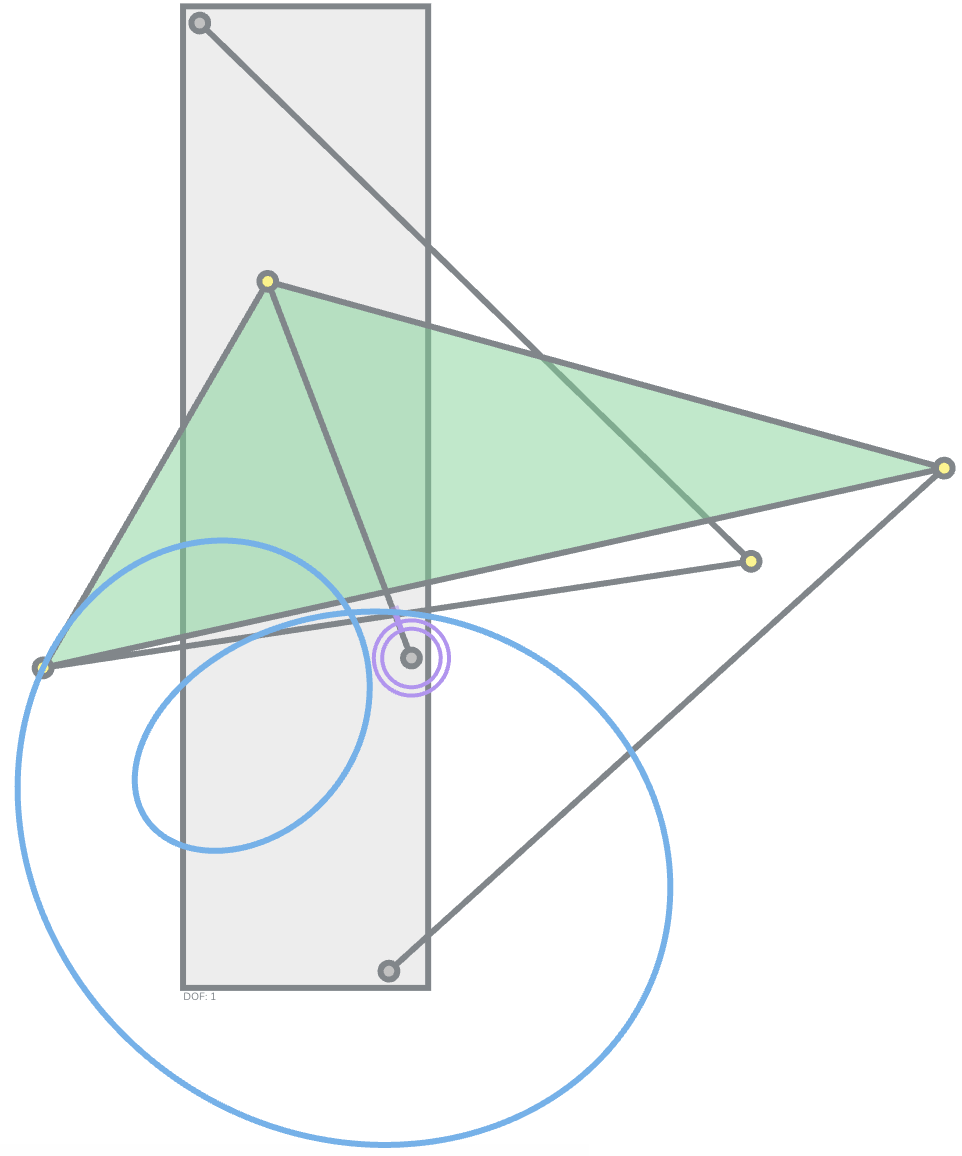} \\[6pt]
\multicolumn{3}{c}{%
\begin{tabular}{@{}c@{\hspace{6pt}}c@{}}
\includegraphics[width=0.31\textwidth]{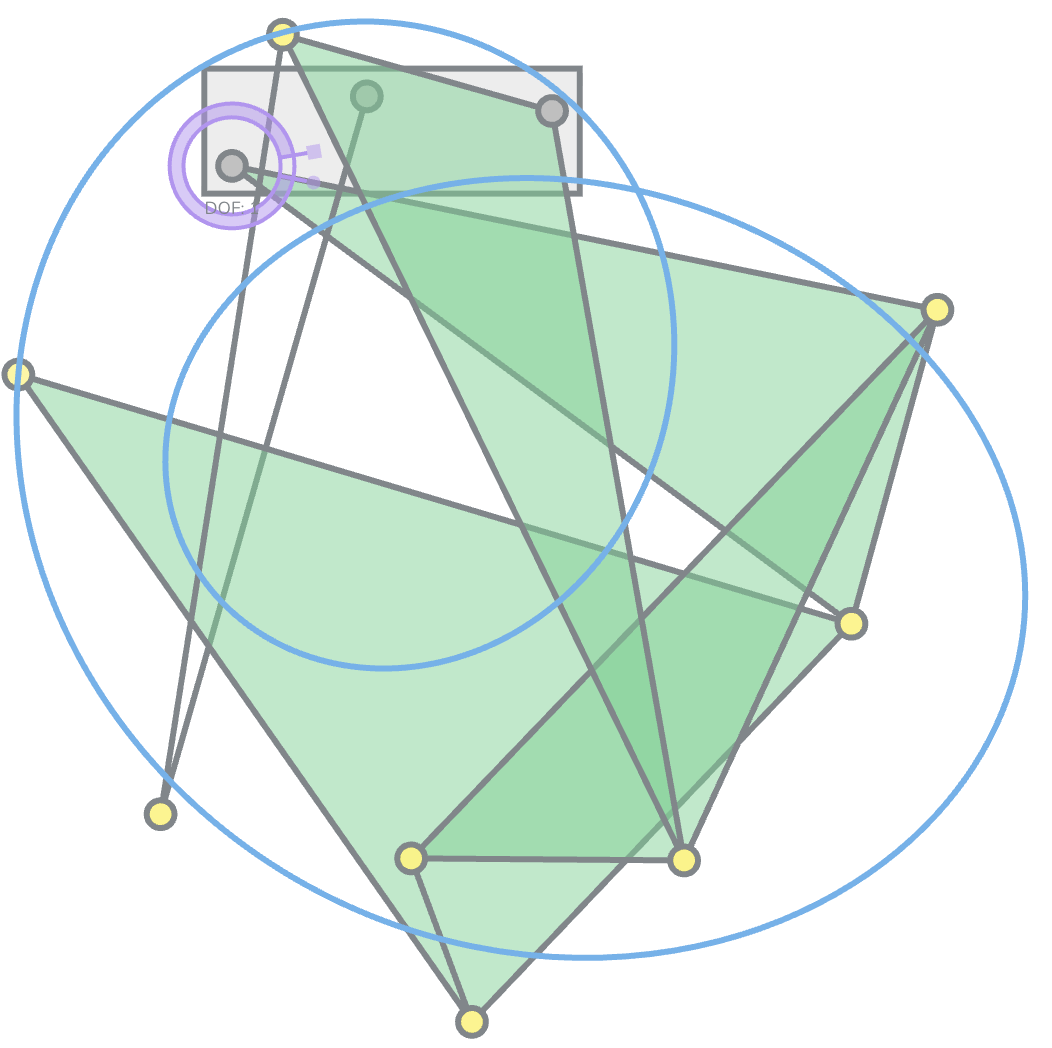} &
\includegraphics[width=0.31\textwidth]{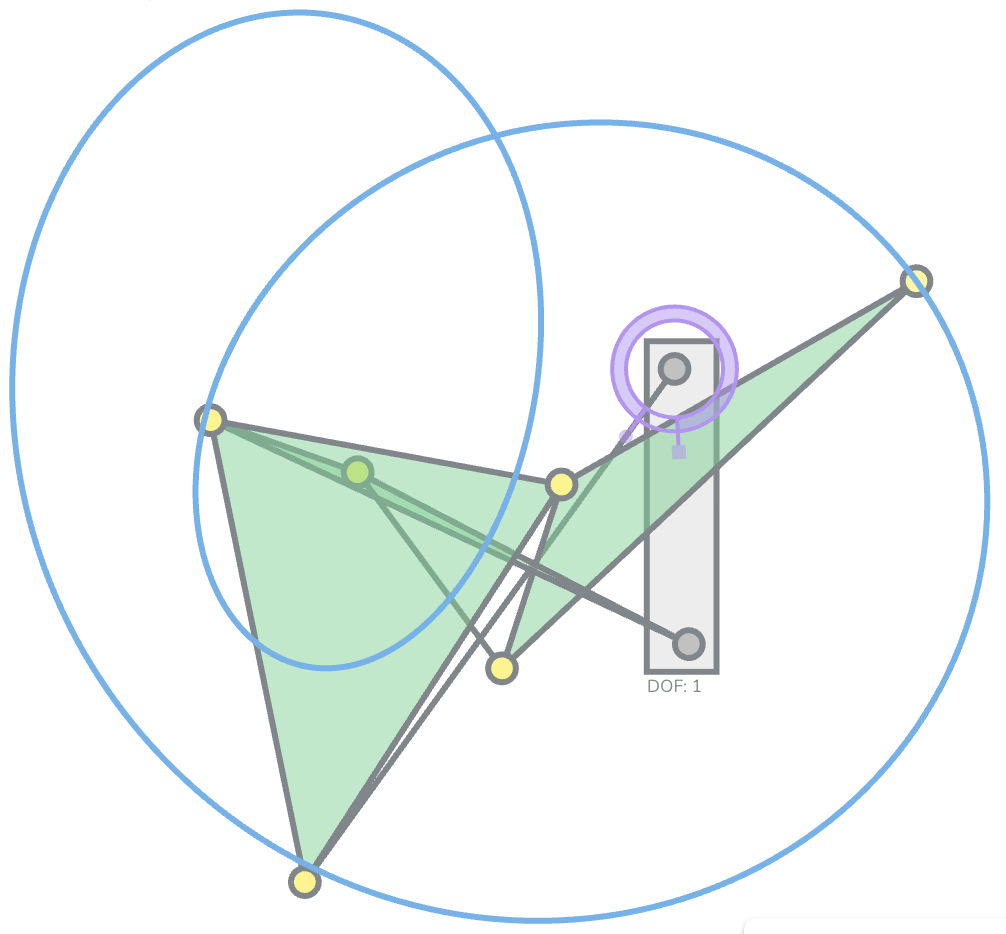} \\
\end{tabular}} \\
\end{tabular}
\caption{Latent KNN baseline: five training-set neighbor latents for a single query coupler curve, each decoded with the same autoregressive transformer and simulated. Each panel shows the generated mechanism layout and the resulting coupler trace relative to the query geometry.}
\label{fig:knn_neighbor_panels}
\end{figure*}

In the second setting, \emph{cross-topology KNN evaluation}, the same 30 neighbor latents for each target curve are decoded under all available mechanism-type conditionings. Specifically, for each neighbor latent, mechanisms are generated using all 33 mechanism families, resulting in $100 \times 30 \times 33$ total predictions. This evaluation substantially expands the design space and reflects a more challenging one-to-many synthesis scenario. Table~\ref{tab:knn_alltypes} reports the corresponding DTW statistics. While the mean DTW increases due to the inclusion of less compatible mechanism families, the minimum DTW remains low, demonstrating that for most target curves there exists at least one mechanism type capable of producing a close geometric match. These KNN tables complement the latent-noise sweep: the latter perturbs the \emph{query} latent and ranks \emph{all} types at each level, whereas KNN varies the conditioning latent among training-set neighbors while using the same decoder and simulator pipeline. Figure~\ref{fig:knn_neighbor_panels} illustrates five representative neighbor-conditioning panels for one query curve.

\begin{figure*}[t]
\centering
\includegraphics[width=\textwidth]{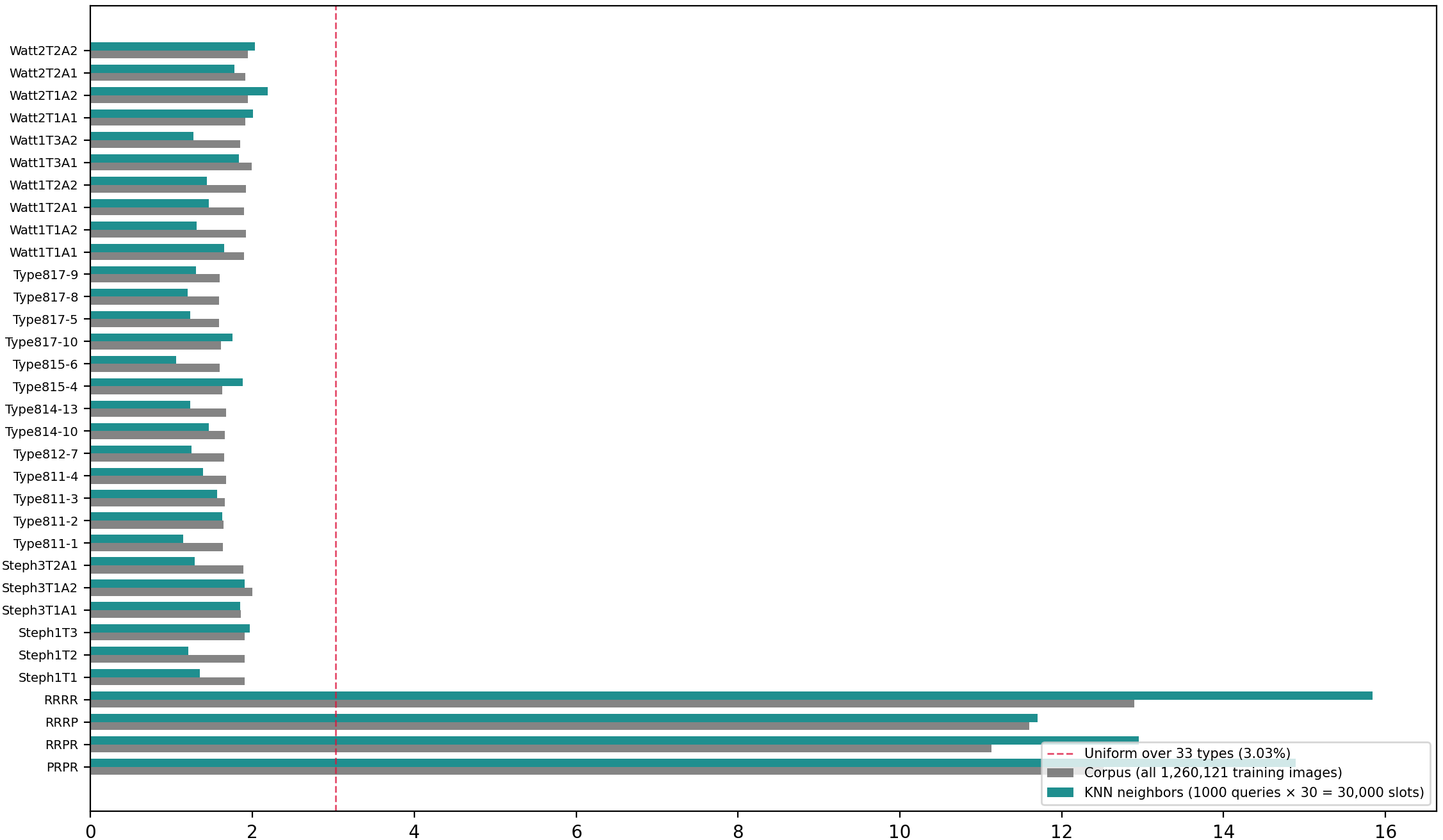}
\caption{Mechanism-type mix: corpus marginal (all $1{,}260{,}121$ training images) vs.\ labels of $30$ VAE nearest neighbors for each of $1{,}000$ random queries ($30{,}000$ neighbor slots). Bar heights are percentages of each pool; dashed line is uniform $1/33$. Pearson $r > 0.99$ between the two percentage profiles. No transformer decoding or curve simulation.}
\label{fig:knn_vae_neighbor_label_hist}
\end{figure*}

Figure~\ref{fig:knn_vae_neighbor_label_hist} compares, on the same axes, two empirical mechanism-type distributions in percentage of their respective pools: (i) the \emph{corpus} marginal from all $1{,}260{,}121$ training coupler-curve images (encoded labels in the released dataset), and (ii) the mix of ground-truth mechanism types among VAE-space $k$-nearest neighbors ($k{=}30$) to $1{,}000$ randomly drawn training latents (Euclidean distance, ball tree over the full corpus; no decoding or simulation). The dashed vertical line marks a uniform $1/33$ share for reference. Neighbor-type percentages track the corpus closely: Pearson correlation between the two $33$-vectors of percentages exceeds $0.99$, so frequent families in the dataset also appear proportionally more often among latent neighbors---as expected if neighborhoods sample the training manifold rather than a single class. The script \texttt{knn\_latent\_neighbor\_label\_histogram.py} regenerates the figure in under one minute from \texttt{vae\_mu.npy} and \texttt{encoded\_labels.npy}.

\begin{table}[t]
\centering
\caption{Latent-space KNN evaluation under matched-topology conditions. Each of the 100 target coupler curves is evaluated using its 30 nearest neighbors in latent space, decoded with the same mechanism type as the target curve, resulting in 3{,}000 total predictions.}
\label{tab:knn_matched}
\begin{tabular}{lcc}
\hline
Metric & Mean & Median \\
\hline
Chamfer Distance (CD) & 0.0071 & 0.0022 \\
Dynamic Time Warping (DTW) & 0.1171 & 0.0030 \\
\hline
\end{tabular}
\end{table}

\begin{table}[t]
\centering
\caption{Latent-space KNN evaluation across all mechanism families. For each of the 100 target coupler curves and its 30 nearest latent neighbors, mechanisms are generated using all 33 mechanism types, resulting in $100 \times 30 \times 33$ total predictions.}
\label{tab:knn_alltypes}
\begin{tabular}{lc}
\hline
Metric & Value \\
\hline
Mean DTW & 0.2288 \\
Minimum DTW & 0.0213 \\
\hline
\end{tabular}
\end{table}

\begin{figure*}[t]
    \centering
    \includegraphics[width=\textwidth]{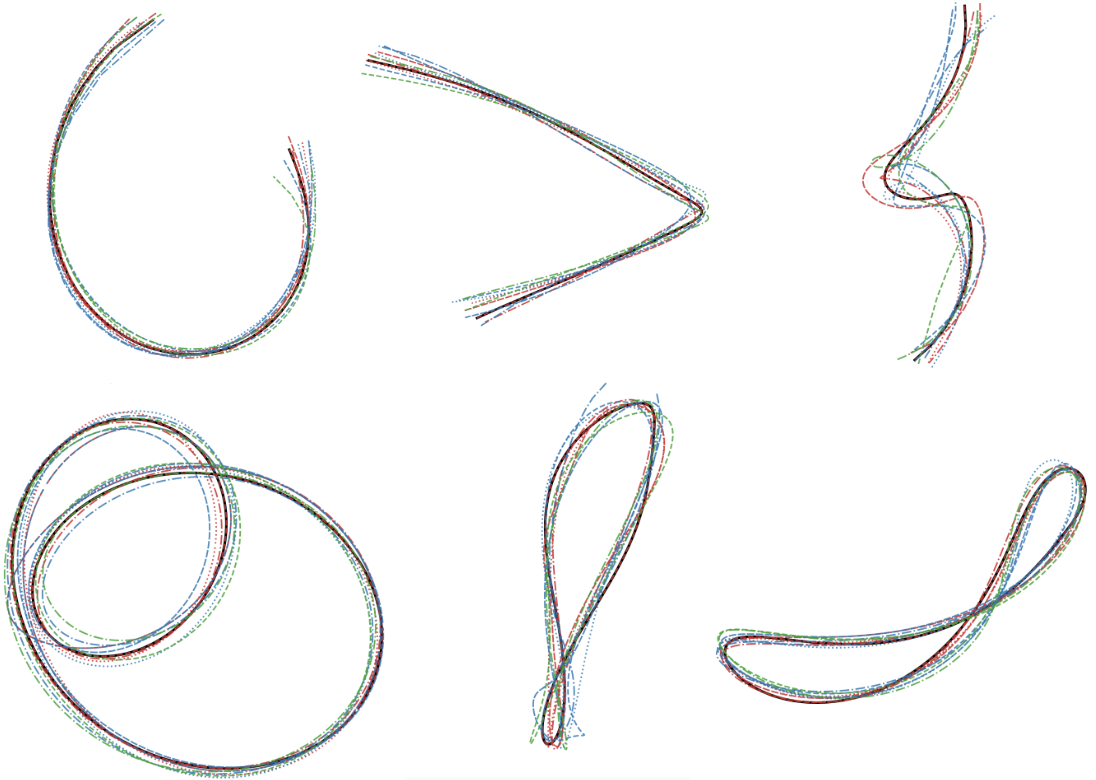}
    \caption{
    Top-10 predicted coupler curves for a single target trajectory.
    The black curve shows the ground-truth coupler curve, while colored dashed curves correspond to the ten best predicted mechanisms ranked by Chamfer Distance after geometric alignment.
    Predictions are drawn from the latent-noise and cross-topology evaluation and from nearest-neighbor latent decoding, illustrating one-to-many synthesis under forward kinematic simulation.
    }
    \label{fig:top10_predictions}
\end{figure*}

Figure~\ref{fig:top10_predictions} compresses the one-to-many behavior already established above: ranked predictions combine the cross-topology latent-noise sweep (including nearest-neighbor latent conditioning in the same simulator metric) and illustrate that multiple linkage layouts can approximate the same target coupler curve after alignment.

In summary, Tables~\ref{tab:latent_noise_gt_mech}--\ref{tab:latent_noise_aggregate} quantify the latent-noise protocol; Tables~\ref{tab:knn_matched}--\ref{tab:knn_alltypes} isolate how conditioning on training-set neighbor latents in VAE space shifts CD/DTW relative to query and noised latents under the \emph{same} decoder and alignment code~\cite{Barrow1977Chamfer,SakoeChiba1978DTW}. The headline aggregate means in Table~\ref{tab:latent_noise_aggregate} match the state-of-the-art claim in the comparison paragraph above; all reported mechanisms are produced by autoregressive decoding~\cite{radford2019language}, not by reading parameters from dataset tables.

\section{Conclusion}
\label{sec:conclusion}

This paper presented a discrete autoregressive transformer for generative mechanism synthesis, framing the inverse mapping from coupler curves to linkage parameters as conditional sequence modeling on a large multi-topology subset of~\cite{Nurizada2025Dataset}. Dual conditioning combines a VAE-derived curve latent with a mechanism-type token (Fig.~\ref{fig:discrete_model}); training adds a geometry-aware bin loss to standard cross-entropy.

The experiments in Section~\ref{sec:results} summarize greedy reconstruction, latent-noise exploration with cross-topology top-$K$ retention, and KNN latent baselines under one simulator-grounded CD/DTW pipeline. Discretization, paired with the proposed conditioning and losses, does not prevent high geometric fidelity under forward simulation.

Future work includes ablations of the hybrid loss and bin resolution~\cite{szegedy2016rethinking,oord2017neural}, alternative noise schedules and candidate counts~$K$, reporting validity rates for simulated assemblies~\cite{craig2005introduction}, richer geometric or kinematic metrics, variable-length decoders~\cite{vaswani2017attention}, explicit constraints during generation, and integration with optimization or human-in-the-loop refinement.



\section*{Data availability}
Mechanism and coupler-curve data are described in prior work (\cite{Nurizada2025Dataset}); evaluation scripts and model checkpoints will be released upon acceptance subject to licensing.

\section*{Declaration of generative AI and AI-assisted technologies in the manuscript preparation process}

During the preparation of this work the author(s) used ChatGPT (OpenAI) and Cursor (AI-assisted editing) in order to proofread the manuscript, revise wording, and improve clarity and coherence of the prose. After using these tools, the author(s) reviewed and edited the content as needed and take(s) full responsibility for the content of the published article.

\bibliographystyle{elsarticle-harv}
\bibliography{eaai}

@article{Nurizada2025Dataset,
  author    = {Anar Nurizada and Rohit Dhaipule and Zhijie Lyu and Anurag Purwar},
  title     = {A Dataset of 3M Single-DOF Planar 4-, 6-, and 8-Bar Linkage Mechanisms With Open and Closed Coupler Curves for Machine Learning-Driven Path Synthesis},
  journal   = {ASME Journal of Mechanical Design},
  volume    = {147},
  number    = {4},
  pages     = {041702},
  year      = {2025},
  publisher = {ASME},
  doi       = {10.1115/1.4067014},
  url       = {https://doi.org/10.1115/1.4067014}
}

@misc{Bolanos2025,
  title         = {MechaFormer: Sequence Learning for Kinematic Mechanism Design Automation},
  author        = {Bolanos, Diana and Ataei, Mohammadmehdi and Jayaraman, Pradeep Kumar},
  year          = {2025},
  eprint        = {2508.09005},
  archivePrefix = {arXiv},
  primaryClass  = {cs.LG},
  howpublished  = {arXiv preprint},
  url           = {https://arxiv.org/abs/2508.09005}
}

@inproceedings{Fan2017Point,
  author    = {Haoqiang Fan and Hao Su and Leonidas J. Guibas},
  title     = {A Point Set Generation Network for 3D Object Reconstruction from a Single Image},
  booktitle = {Proceedings of the IEEE Conference on Computer Vision and Pattern Recognition (CVPR)},
  year      = {2017},
  pages     = {2463--2472}
}

@article{Nurizada2025cBetaVAE,
  author    = {Anar Nurizada and Zhijie Lyu and Anurag Purwar},
  title     = {Path Generative Model Based on Conditional {$\beta$}-Variational Auto Encoder for Four-Bar Mechanism Design},
  journal   = {Journal of Mechanisms and Robotics},
  volume    = {17},
  number    = {6},
  pages     = {061004},
  year      = {2025},
  publisher = {ASME},
  doi       = {10.1115/1.40637026},
  url       = {https://asmedigitalcollection.asme.org/mechanismsrobotics/article/17/6/061004/1209221}
}

@misc{nobari2024linklearningjointrepresentations,
  title         = {LInK: Learning Joint Representations of Design and Performance Spaces through Contrastive Learning for Mechanism Synthesis},
  author        = {Nobari, Amin Heyrani and Srivastava, Akash and Gutfreund, Dan and Xu, Kai and Ahmed, Faez},
  year          = {2024},
  eprint        = {2405.20592},
  archivePrefix = {arXiv},
  primaryClass  = {cs.LG},
  howpublished  = {arXiv preprint},
  url           = {https://arxiv.org/abs/2405.20592}
}

@article{Nurizada2024InvariantCoupler,
  author    = {Anar Nurizada and Anurag Purwar},
  title     = {An Invariant Representation of Coupler Curves Using a Variational Autoencoder: Application to Path Synthesis of Four-Bar Mechanisms},
  journal   = {Journal of Computing and Information Science in Engineering},
  volume    = {24},
  number    = {1},
  pages     = {011008},
  year      = {2024},
  publisher = {ASME},
  doi       = {10.1115/1.4063726},
  url       = {https://doi.org/10.1115/1.4063726}
}

@inproceedings{Nobari2022LINKSDataset,
  author       = {Amin Heyrani Nobari and Akash Srivastava and Dan Gutfreund and Faez Ahmed},
  title        = {LINKS: A Dataset of a Hundred Million Planar Linkage Mechanisms for Data-Driven Kinematic Design},
  booktitle    = {Proceedings of the ASME 2022 International Design Engineering Technical Conferences \& Computers and Information in Engineering Conference (IDETC/CIE2022)},
  series       = {Volume 3A: Advanced Engineering Informatics and Digital Product Development},
  pages        = {V03AT03A013},
  year         = {2022},
  organization = {ASME},
  doi          = {10.1115/DETC2022-89798},
  url          = {https://doi.org/10.1115/DETC2022-89798}
}

@article{touvron2023llama,
  title         = {{LLaMA}: Open and Efficient Foundation Language Models},
  author        = {Touvron, Hugo and Lavril, Thibaut and Izacard, Gautier and others},
  year          = {2023},
  eprint        = {2302.13971},
  archivePrefix = {arXiv},
  primaryClass  = {cs.CL}
}

@inproceedings{zhang2019root,
  title     = {Root Mean Square Layer Normalization},
  author    = {Zhang, Biao and Sennrich, Rico},
  booktitle = {Advances in Neural Information Processing Systems},
  volume    = {32},
  year      = {2019}
}

@inproceedings{vaswani2017attention,
  title     = {Attention Is All You Need},
  author    = {Vaswani, Ashish and Shazeer, Noam and Parmar, Niki and others},
  booktitle = {Advances in Neural Information Processing Systems},
  volume    = {30},
  year      = {2017}
}

@article{ramachandran2017searching,
  title   = {Searching for Activation Functions},
  author  = {Ramachandran, Prajit and Zoph, Barret and Le, Quoc V.},
  journal = {arXiv preprint arXiv:1710.05941},
  year    = {2017}
}

@article{williams1989learning,
  title   = {A Learning Algorithm for Continually Running Fully Recurrent Neural Networks},
  author  = {Williams, Ronald J. and Zipser, David},
  journal = {Neural Computation},
  volume  = {1},
  number  = {2},
  pages   = {270--280},
  year    = {1989}
}

@article{Wampler1992,
  author  = {Wampler, Charles W. and Morgan, Alexander P. and Sommese, Andrew J.},
  title   = {Complete Real Solution in $(9,9)$ for Nine-Point Path Synthesis for Four-Bars},
  journal = {Journal of Mechanical Design},
  year    = {1992},
  volume  = {114},
  number  = {1},
  pages   = {153--159},
  doi     = {10.1115/1.2916908},
}

@Article{Bai2015,
  author    = {Shaoping Bai and Jorge Angeles},
  journal   = {Mechanism and Machine Theory},
  title     = {Coupler-curve synthesis of four-bar linkages via a novel formulation},
  year      = {2015},
  month     = {dec},
  pages     = {177--187},
  volume    = {94},
  doi       = {10.1016/j.mechmachtheory.2015.08.010},
  publisher = {Elsevier {BV}},
}

@InProceedings{Vermeer2018,
  author    = {Vermeer, Kaz and Kuppens, Reinier and Herder, Justus},
  booktitle = {Volume 2A: 44th Design Automation Conference},
  title     = {Kinematic Synthesis Using Reinforcement Learning},
  year      = {2018},
  month     = {aug},
  pages     = {V02AT03A030},
  publisher = {American Society of Mechanical Engineers},
  doi       = {10.1115/detc2018-85529}
}

@Article{Fogelson2023,
  author    = {Fogelson, Mitchell B. and Tucker, Conrad and Cagan, Jonathan},
  journal   = {Journal of Mechanical Design},
  title     = {{GCP}-{HOLO}: Generating High-Order Linkage Graphs for Path Synthesis},
  year      = {2023},
  month     = {apr},
  number    = {7},
  volume    = {145},
  pages     = {071704},
  doi       = {10.1115/1.4062147},
  publisher = {ASME International}
}

@inproceedings{Barrow1977Chamfer,
  author    = {Barrow, Hugh G. and Tenenbaum, Jay M. and Hanson, Allen R. and Selfridge, Peter J.},
  title     = {Parametric Correspondence and Chamfer Matching: Two New Techniques for Image Matching},
  booktitle = {Proceedings of the 5th International Joint Conference on Artificial Intelligence (IJCAI)},
  year      = {1977},
  pages     = {659--663},
}

@article{SakoeChiba1978DTW,
  author  = {Sakoe, Hiroaki and Chiba, Seibi},
  title   = {Dynamic Programming Algorithm Optimization for Spoken Word Recognition},
  journal = {IEEE Transactions on Acoustics, Speech, and Signal Processing},
  year    = {1978},
  volume  = {26},
  number  = {1},
  pages   = {43--49},
  doi     = {10.1109/TASSP.1978.1163055},
}

@InProceedings{Hoskins1993,
  author    = {Hoskins, J.C. and Kramer, G.A.},
  booktitle = {IEEE International Conference on Neural Networks},
  title     = {Synthesis of mechanical linkages using artificial neural networks and optimization},
  year      = {1993},
  pages     = {822-J},
  doi       = {10.1109/ICNN.1993.298663},
}

@article{vasiliu2001,
   author = {Vasiliu, A. and Yannou, B.},
   title = {Dimensional synthesis of planar mechanisms using neural networks: application to path generator linkages},
   journal = {Mechanism and Machine Theory},
   volume = {36},
   number = {2},
   pages = {299-310},
   year = {2001}
}

@inproceedings{Xie2007,
  author    = {Xie, J. and Chen, Y.},
  title     = {Application of Back Propagation Neural Network to Synthesis of Whole Cycle Motion Generation Mechanism},
  booktitle = {Proceedings of the 12th {IFToMM} World Congress},
  address   = {Besan\c{c}on, France},
  year      = {2007},
  pages     = {17--21},
}

@article{galan2009,
   author = {Galan-Marin, G. and Alonso, F. J. and Del Castillo, J. M.},
   title = {Shape optimization for path synthesis of crank-rocker mechanisms using a wavelet-based neural network},
   journal = {Mechanism and Machine Theory},
   volume = {44},
   number = {6},
   pages = {1132-1143},
   year = {2009}
}

@article{khan2015,
    author = {Khan, N. and Ullah, I. and Al-Grafi, M.},
    title = {Dimensional synthesis of mechanical linkages using artificial neural networks and Fourier descriptors},
    journal = {Mechanical Sciences},
    volume = {6},
    number = {1},
    pages = {29-34},
    year = {2015}
 }

@Article{Ahmadi2016,
  author    = {Ahmadi, Bahman and Nariman-zadeh, Nader and Jamali, Ali},
  journal   = {Engineering Optimization},
  title     = {Path synthesis of four-bar mechanisms using synergy of polynomial neural network and Stackelberg game theory},
  year      = {2016},
  issn      = {1029-0273},
  month     = aug,
  number    = {6},
  pages     = {932--947},
  volume    = {49},
  doi       = {10.1080/0305215x.2016.1218641},
  publisher = {Informa UK Limited},
}

@Article{Li2017,
  author    = {Li, Xiangyun and Chen, Peng},
  journal   = {Mathematical Problems in Engineering},
  title     = {A Parametrization-Invariant Fourier Approach to Planar Linkage Synthesis for Path Generation},
  year      = {2017},
  issn      = {1563-5147},
  pages     = {1--16},
  volume    = {2017},
  doi       = {10.1155/2017/8458149},
  publisher = {Hindawi Limited},
}

@InBook{Mo2019,
  author    = {Mo, Xiaojuan and Ge, Wenjie and Zhao, Donglai and Zhang, Yaqing},
  pages     = {32--41},
  publisher = {Springer Singapore},
  title     = {Path Synthesis of Crank-Rocker Mechanism Using Fourier Descriptors Based Neural Network},
  year      = {2019},
  isbn      = {9789811501425},
  month     = sep,
  booktitle = {Recent Advances in Mechanisms, Transmissions and Applications},
  doi       = {10.1007/978-981-15-0142-5_4},
  issn      = {2211-0992},
}

@Article{Yim2021,
  author    = {Yim, Neung Hwan and Lee, Jongjun and Kim, Jungho and Kim, Yoon Young},
  journal   = {Mechanism and Machine Theory},
  title     = {Big data approach for the simultaneous determination of the topology and end-effector location of a planar linkage mechanism},
  year      = {2021},
  issn      = {0094-114X},
  month     = sep,
  pages     = {104375},
  volume    = {163},
  doi       = {10.1016/j.mechmachtheory.2021.104375},
  publisher = {Elsevier BV},
}

@Article{Kapsalyamov2023,
  author    = {Akim Kapsalyamov and Shahid Hussain and Nicholas A.T. Brown and Roland Goecke and Munawar Hayat and Prashant K. Jamwal},
  journal   = {Engineering Applications of Artificial Intelligence},
  title     = {Synthesis of a six-bar mechanism for generating knee and ankle motion trajectories using deep generative neural network},
  year      = {2023},
  month     = {jan},
  pages     = {105500},
  volume    = {117},
  doi       = {10.1016/j.engappai.2022.105500},
  publisher = {Elsevier {BV}},
}

@InProceedings{Yim2023,
  author    = {Yim, Neung Hwan and Ryu, Jegyeong and Kim, Yoon Young},
  booktitle = {2023 IEEE International Conference on Robotics and Automation (ICRA)},
  title     = {Big data approach for synthesizing a spatial linkage mechanism},
  year      = {2023},
  pages     = {7433-7439},
  doi       = {10.1109/ICRA48891.2023.10161300},
}

@inproceedings{kingma2014autoencoding,
  title     = {Auto-Encoding Variational {B}ayes},
  author    = {Kingma, Diederik P. and Welling, Max},
  booktitle = {International Conference on Learning Representations},
  year      = {2014},
  note      = {ICLR 2014}
}

@inproceedings{loshchilov2017adamw,
  title     = {Decoupled Weight Decay Regularization},
  author    = {Loshchilov, Ilya and Hutter, Frank},
  booktitle = {International Conference on Learning Representations},
  year      = {2019}
}

@inproceedings{szegedy2016rethinking,
  title     = {Rethinking the {I}nception Architecture for Computer Vision},
  author    = {Szegedy, Christian and Vanhoucke, Vincent and Ioffe, Sergey and Shlens, Jon and Wojna, Zbigniew},
  booktitle = {Proceedings of the {IEEE} Conference on Computer Vision and Pattern Recognition},
  pages     = {2818--2826},
  year      = {2016}
}

@inproceedings{radford2021learning,
  title     = {Learning Transferable Visual Models From Natural Language Supervision},
  author    = {Radford, Alec and Kim, Jong Wook and Hallacy, Chris and Ramesh, Aditya and Goh, Gabriel and Agarwal, Sandhini and Sastry, Girish and Askell, Amanda and Mishkin, Pamela and Clark, Jack and Krueger, Gretchen and Sutskever, Ilya},
  booktitle = {Proceedings of the 38th International Conference on Machine Learning},
  pages     = {8748--8763},
  year      = {2021},
  volume    = {139},
  series    = {Proceedings of Machine Learning Research},
  publisher = {PMLR}
}

@inproceedings{BerndtClifford1994DTW,
  author    = {Berndt, Donald J. and Clifford, James},
  title     = {Using Dynamic Time Warping to Find Patterns in Time Series},
  booktitle = {Proceedings of the {AAAI} Workshop on Knowledge Discovery in Databases},
  year      = {1994},
  pages     = {359--370}
}

@book{Erdman2001MechanismDesign,
  author    = {Erdman, Arthur G. and Sandor, George N. and Kota, Sridhar},
  title     = {Mechanism Design: Volume 1, Analysis},
  edition   = {4},
  publisher = {Prentice Hall},
  address   = {Upper Saddle River, NJ},
  year      = {2001}
}

@techreport{radford2019language,
  title       = {Language Models are Unsupervised Multitask Learners},
  author      = {Radford, Alec and Wu, Jeffrey and Child, Rewon and Luan, David and Amodei, Dario and Sutskever, Ilya},
  institution = {OpenAI},
  year        = {2019}
}

@inproceedings{sutskever2014sequence,
  title     = {Sequence to Sequence Learning with Neural Networks},
  author    = {Sutskever, Ilya and Vinyals, Oriol and Le, Quoc V.},
  booktitle = {Advances in Neural Information Processing Systems},
  volume    = {27},
  year      = {2014}
}

@inproceedings{He2016ResNet,
  title     = {Deep Residual Learning for Image Recognition},
  author    = {He, Kaiming and Zhang, Xiangyu and Ren, Shaoqing and Sun, Jian},
  booktitle = {Proceedings of the {IEEE} Conference on Computer Vision and Pattern Recognition},
  pages     = {770--778},
  year      = {2016}
}

@inproceedings{oord2017neural,
  title     = {Neural Discrete Representation Learning},
  author    = {van den Oord, Aaron and Vinyals, Oriol and Kavukcuoglu, Koray},
  booktitle = {Advances in Neural Information Processing Systems},
  volume    = {30},
  year      = {2017}
}

@book{craig2005introduction,
  title     = {Introduction to Robotics: Mechanics and Control},
  author    = {Craig, John J.},
  edition   = {3},
  publisher = {Pearson Prentice Hall},
  address   = {Upper Saddle River, NJ},
  year      = {2005}
}

@article{cover1967nearest,
  title   = {Nearest Neighbor Pattern Classification},
  author  = {Cover, Thomas and Hart, Peter},
  journal = {IEEE Transactions on Information Theory},
  volume  = {13},
  number  = {1},
  pages   = {21--27},
  year    = {1967}
}

@inproceedings{higgins2017beta,
  title     = {{$\beta$}-{VAE}: Learning Basic Visual Concepts with a Constrained Variational Framework},
  author    = {Higgins, Irina and Matthey, Loic and Pal, Arka and Burgess, Christopher and Glorot, Xavier and Botvinick, Matthew and Mohamed, Shakir and Lerchner, Alexander},
  booktitle = {International Conference on Learning Representations},
  year      = {2017},
  note      = {ICLR 2017}
}

@article{shannon1948mathematical,
  title   = {A Mathematical Theory of Communication},
  author  = {Shannon, Claude E.},
  journal = {Bell System Technical Journal},
  volume  = {27},
  number  = {3},
  pages   = {379--423},
  year    = {1948}
}

@book{nocedal2006numerical,
  title     = {Numerical Optimization},
  author    = {Nocedal, Jorge and Wright, Stephen J.},
  edition   = {2},
  publisher = {Springer},
  address   = {New York},
  year      = {2006}
}

@inproceedings{paszke2019pytorch,
  title     = {{PyTorch}: An Imperative Style, High-Performance Deep Learning Library},
  author    = {Paszke, Adam and Gross, Sam and Massa, Francisco and Lerer, Adam and Bradbury, James and Chanan, Gregory and Killeen, Trevor and Lin, Zeming and Gimelshein, Natalia and Antiga, Luca and Desmaison, Alban and Kopf, Andreas and Yang, Edward and DeVito, Zach and Raison, Martin and Tejani, Alykhan and Chilamkurthy, Sasank and Steiner, Benoit and Fang, Lu and Bai, Junjie and Chintala, Soumith},
  booktitle = {Advances in Neural Information Processing Systems},
  volume    = {32},
  year      = {2019}
}

@book{docarmo1976differential,
  title     = {Differential Geometry of Curves and Surfaces},
  author    = {do Carmo, Manfredo P.},
  publisher = {Prentice-Hall},
  address   = {Englewood Cliffs, NJ},
  year      = {1976}
}

@article{Tang2026,
  author  = {Tang, Ray and Lyu, Zhijie and Purwar, Anurag},
  title   = {Kinematic Synthesis of Planar Leg Mechanisms Through Large-Scale Dataset Generation, Geometric Filtering, and Optimization},
  journal = {Machines},
  volume  = {14},
  number  = {3},
  pages   = {253},
  year    = {2026},
  url     = {https://www.mdpi.com/2075-1702/14/3/253}
}

\end{document}